\documentclass{article} 
\usepackage{nips11submit_e_mod,times}

\pdfoutput=1

\usepackage{times}
\usepackage{epsfig}
\usepackage{graphicx}
\usepackage{amsmath}
\usepackage{amssymb}

\usepackage{fancyhdr}

\usepackage{array}

\title{Early Recognition of Human Activities from First-Person Videos Using Onset Representations}

\author{M. S. Ryoo$^{1}$, Thomas J. Fuchs$^{1}$, Lu Xia$^{2}$, J. K. Aggarwal$^{2}$, and Larry Matthies$^{1}$ \\
$^{1}$Jet Propulsion Laboratory, California Institute of Technology, Pasadena, CA, U.S.A.\\
$^{2}$Department of ECE, The University of Texas at Austin, Austin, TX, U.S.A.\\
\texttt{mryoo@jpl.nasa.gov} \\
}

\nipsfinalcopy

\begin{document}

\maketitle

\pagestyle{fancy}
\fancyhf{}
\lhead{
\color{blue}
The final version of this manuscript was published at HRI 2015. The new version contains polished equations and better descriptions, so please read/cite the revised version instead:\\
\color{black}
M. S. Ryoo, T. J. Fuchs, L. Xia, J. K. Aggarwal, and L. Matthies, ``Robot-Centric Activity Prediction from First-Person Videos: What Will They Do to Me?'', ACM/IEEE International Conference on Human-Robot Interaction (HRI), 2015.
}
\renewcommand{\headrulewidth}{0pt}
\renewcommand{\headrule}{{\hrule width\headwidth height\headrulewidth \vskip-25pt}}

\begin{abstract}
In this paper, we propose a methodology for \emph{early recognition} of human activities from videos taken with a first-person viewpoint. Early recognition, which is also known as activity prediction, is an ability to infer an ongoing activity at its early stage. We present an algorithm to perform recognition of activities targeted at the camera from streaming videos, making the system to predict intended activities of the interacting person and avoid harmful events before they actually happen. We introduce the novel concept of `onset' that efficiently summarizes pre-activity observations, and design an approach to consider event history in addition to ongoing video observation for early first-person recognition of activities. We propose to represent onset using cascade histograms of time series gradients, and we describe a novel algorithmic setup to take advantage of onset for early recognition of activities. The experimental results clearly illustrate that the proposed concept of onset enables better/earlier recognition of human activities from first-person videos.

\end{abstract}



\section{Introduction}

First-person activity recognition is a research area studying automated recognition of human activities from videos with the actor's own viewpoint. Its main difference to the conventional 3rd-person activity recognition is that the observer wearing the camera himself/herself (e.g., a robot) is involved in the ongoing activity, making its perception to become egocentric videos. Particularly, a recent research work \cite{ryoo13} demonstrated recognition approaches for interaction-level human activities from first-person videos, which illustrated its potential for real-world real-time applications including assistive wearable systems and intelligent robotics.


An ability particularly important and necessary for first-person recognition systems is the ability to infer humans' intended activities at their early stage. A wearable system must recognize ongoing events around the human as early as possible to provide appropriate service for human tasks and to alarm accidents like `a car running into the person'. Similarly, public service robots and surveillance/military robots must protect themselves from any harmful events by inferring the beginning of dangerous activities like an `assault'. Natural human-robot interaction also becomes possible by providing early reaction to humans' actions. This is not just about real-time implementations of activity recognition, but more about recognition of activities from observations only containing the beginning part of the activity. The objective is to detect an ongoing activity in the middle of the activity execution, before it is completed.

This problem, recognition of an activity before fully observing its execution, is called `early recognition' or `activity prediction' \cite{ryoo11}. However, even though there are recent works on early recognition, (1) it has never been studied for first-person videos and (2) research on early recognition approach that simultaneously considers pre-activity observations and ongoing activity videos has been limited. In real-world first-person recognition scenarios, the system is required to continuously process long video inputs containing a sequence of multiple activities. As a consequence, it becomes necessary to consider the video stream of the ongoing activity (e.g., a video segment corresponding to the first half of the activity) as well as history or signals observed prior to the activity. In this paper, we call such signals observed before the activity as \emph{onsets} of the activity.



This paper newly introduces the concept of onsets, and presents an early recognition approach to take advantage of them. We formulate the early recognition (i.e., prediction) problem to consider activity history and human intention together with ongoing observation of the activity, and discuss how our \emph{onset signatures} enable abstraction of such pre-activity observations for better recognition of activities. We define an onset activity as short and subtle human motion (e.g., waving and reaching) observable before main activities (e.g., shaking hands and throwing an object), and attempt to capture/model onset patterns displayed prior to each main activity. More specifically, we compute a collection of weak classifier responses (each corresponding to a particular onset) over time and construct cascade histograms of their time series gradients as our representation summarizing pre-activity observations: onset signatures. Our method is particularly designed to capture loose stochastic correlations between onsets and the target activities (e.g., reaching an object may or may not occur before throwing but they are correlated) and also consider absence of certain onsets (e.g., absence of waving before punching) for better recognition. An efficient (linear time complexity) algorithm is designed to take advantage of our onset signatures to perform better early recognition.



\renewcommand{\headrulewidth}{0.4pt}

\subsection{Related work}
\label{subsec:related}

The research area of first-person activity recognition is gaining an increasing amount of attention recently. There are several works on recognition of ego-actions of the person (i.e., actions of the person wearing a camera such as skiing) \cite{kitani11,fathi11}, object-oriented analysis of humans using objects (e.g., a towel) \cite{lee12,ramanan12}, or analysis based on face and gaze \cite{fathi2012social}. However, only very few works considered recognition of interaction-level activities where multiple humans (or robots) physically interact each other \cite{ryoo13}. 

The problem of early recognition was introduced and formulated with modern spatio-temporal features in \cite{ryoo11}, but it was limited to 3rd-person videos and did not consider pre-activity observations. There also have been works considering past activity history for predicting future states/locations using state-models \cite{saxena13} and/or trajectories \cite{kitani12,xie13} from 3rd-person videos. However, even though these approaches are appropriate for predicting future steps of the activities composed of clear states, they are unsuitable for directly handling dynamic first-person videos whose analysis requires various types of spatio-temporal video features \cite{laptev05,dollar05,corso12} that display highly sparse and noisy characteristics. In order to enable accurate early recognition for interaction-level first-person activities, simultaneous consideration of pre-activity observations (i.e., onsets) and ongoing activity observations is needed. 

To our knowledge, this paper is the first paper is discuss `early recognition' problem for first-person videos. We also believe it is the first paper to explicitly consider \emph{pre-activity observations} (i.e., frames `before' the starting time of the activity) for recognition, which was not attempted in \cite{ryoo11,hoai12}.

\section{Problem formulation}
\label{sec:formulation}

In this section, we introduce the problem of \emph{early detection} of activities and its formulation to consider pre-activity video observations, while also mentioning conventional \emph{activity detection}.

{\flushleft\textbf{Human activity detection:} A typical problem of detecting an activity $C$ given a continuous video $V$ at time frame $t$ can be formulated as:}
\begin{equation}
	\begin{aligned}
		P(C^t ~|~ V) = \sum_{t'} P(C^{[t', t]} ~|~ V)  = \frac{\sum_{t'}P(V[t', t] ~|~ C^{[t', t]}) P(C, [t, t'])}{\sum_{C, t'}{P(V[t', t] ~|~ C^{[t, t']}) P(C, [t, t'])}}
	\end{aligned}
	\label{eq:basic}
\end{equation}
where $[t', t]$ are the time intervals ending with current time $t$ (i.e., $t'$ is possible starting times).

Given continuous video observation $V$, the system only focuses on the video segment $V[t', t]$ corresponding to the activity's time intervals. We call this problem more specifically as `after-the-fact detection'. This must be applied for all possible $[t', t]$ to process the entire video.

\begin{figure}
	\centering
	\begin{minipage}[t][][b]{0.72\linewidth}
		\resizebox{1.0\linewidth}{!}{%
		  \includegraphics{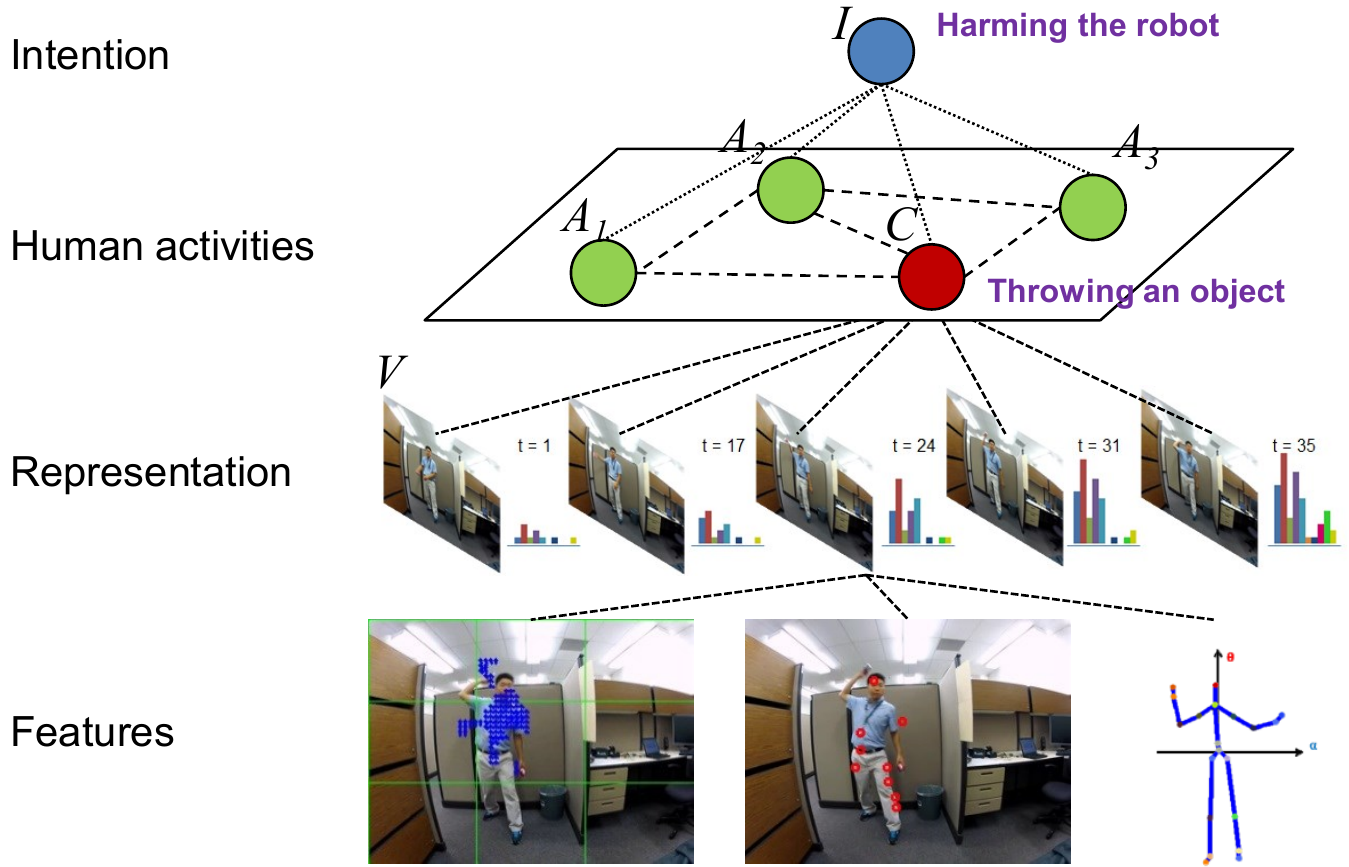}
		}
	\end{minipage}\hfill
	\begin{minipage}[t]{0.22\linewidth}
		\vspace{0pt}
		\caption{Graphical model representation of the scenario where an interacting human performs a sequence of activities under a specific intention. The recognition process is required to consider activity-activity relations, capturing pre-activity observations.}
		\label{fig:model}
	\end{minipage}
\end{figure}

{\flushleft\textbf{Early detection of human activities:} The early detection problem (also called `activity prediction' problem) is the problem of recognizing ongoing activities at their early stage. In contrast to after-the-fact detection, recognition must be made in the middle of the activity before it is fully executed. That is, the system is required to explicitly consider multiple progress levels $d$ of the activity $C$:}
\begin{equation}
	\begin{aligned}
		P(C^t | V) = \frac{\sum_d \sum_{[t_1, t_2]} P(V[t_1, t_2] ~|~ C^{[t_1, t_2]}, d) P(C^{[t_1, t_2]}, d)}{\sum_{C, d, [t_1, t_2]} P(V[t_1, t_2] ~|~ C^{[t_1, t_2]}, d) P(C^{[t_1, t_2]}, d)}
	\end{aligned}
\end{equation}
where $t = t_1 + d \cdot (t_2 - t_1)$. The system must check all candidate time intervals $[t_1, t_2]$ while also considering that the current time $t$ maybe in the middle of the interval and the activity is still ongoing. We call this the early `detection' problem, which extends the early `classification' problem (i.e., early categorization of pre-segmented videos) introduced in \cite{ryoo11}. 


{\flushleft\textbf{Early detection of human activities with context:} Even though the above formulation enables early detection of activities, it is often insufficient for continuous video scenarios. It only utilizes the video segment corresponding to the time interval alone to make the decision, while ignoring all previous video observations. In continuous videos, activities occur in a sequence and they are correlated. Furthermore, the interacting person usually has his/her own intention, such as `harming' the camera or `avoiding' the robot. Figure \ref{fig:model} illustrates a graphical model describing such activity-activity relations and intention-activity relations. Thus, the standard detection problem (i.e., after-the-fact detection) can be formulated as:}
\begin{equation}
	\begin{aligned}
		P(C^{t} ~|~ V) =& \frac{\sum_{([t_1, t_2], \textbf{A}, I)} P(V ~|~ C^{[t_1, t_2]}, \textbf{A}, I)}{\sum_{(C, [t_1,t_2], \textbf{A}, I)} P(V~|~C^{[t_1, t_2]}, \textbf{A}, I)} \propto \sum_{[t_1, t_2], \textbf{A}, I} F(C^{[t_1, t_2]}, \textbf{A}, I, V)
	\end{aligned}
\end{equation}
where $\textbf{A}$ is a set of all previous activities and $I$ is the intention of the interacting person, while assuming a uniform prior. The function $F(C^{[t_{1}, t_{2}]}, \textbf{A}, I, V)$ is what we use to estimate the likelihood measure $P(V~|~C^{[t_1, t_2]}, \textbf{A}, I)$, and it must be designed to consider activity-activity relations (i.e., $\textbf{A} = \{A_1, \cdots, A_{|\textbf{A}|}\}$ and $C$) as well as intention-activity relations displayed in Figure \ref{fig:model}.

The \emph{early detection} problem can be formulated similarly so that it uses pre-activity information, while making the formulation explicitly consider the progress level $d$ of the current activity $C$:
\begin{equation}
	\begin{aligned}
		P(C^t ~|~ V) = \sum_d \sum_{[t_1, t_2]} P(C^{[t_1, t_2]}, d ~|~ V) \propto \sum_d \sum_{[t_1, t_2]} \sum_{(\textbf{A}, I)} F(C^{[t_1, t_2]}, d, \textbf{A}, I, V)
	\end{aligned}
\end{equation}
where $t = t_1 + d \cdot (t_2 - t_1)$.

The key issues for the early detection are (i) designing the robust likelihood function $F$, (ii) learning the model parameters in $F$ from training videos, and (iii) inference given a new video observation $V$. This inference must be made at its every time frame $t$ while considering possible intervals $[t_{1}, t_{2}]$ containing $t$. Notice that $t$ is smaller than $t_2$ in the case of an ongoing activity (i.e., it is in the middle of execution), and $F$ must be designed to consider such characteristic.


{\flushleft\textbf{Challenges:} The main technical challenge we need to face is that the above-mentioned computations must be performed in real-time and the decision whether the activity is occurring or not must be made as early as possible. Particularly in first-person vision applications (e.g., a robot), it is very contradictory to say ``even though the approach is able to recognize activities only using initial parts of their videos, it will take tens of seconds or minutes to process them''. This implies that (1) we need to apply the recognition algorithm almost every frame (i.e., it should not wait) and that (2) the algorithm still must perform in real-time or faster. This makes a standard way of modeling/training the likelihood function $F$ and making an multi-iteration inference using a latent SVM formulation similar to \cite{lan12} or MCMC-based searching difficult.}



\section{Early activity detection using onsets}
\label{sec:approach}

\begin{figure}
	\centering
	\resizebox{0.89\linewidth}{!}{%
	  \includegraphics{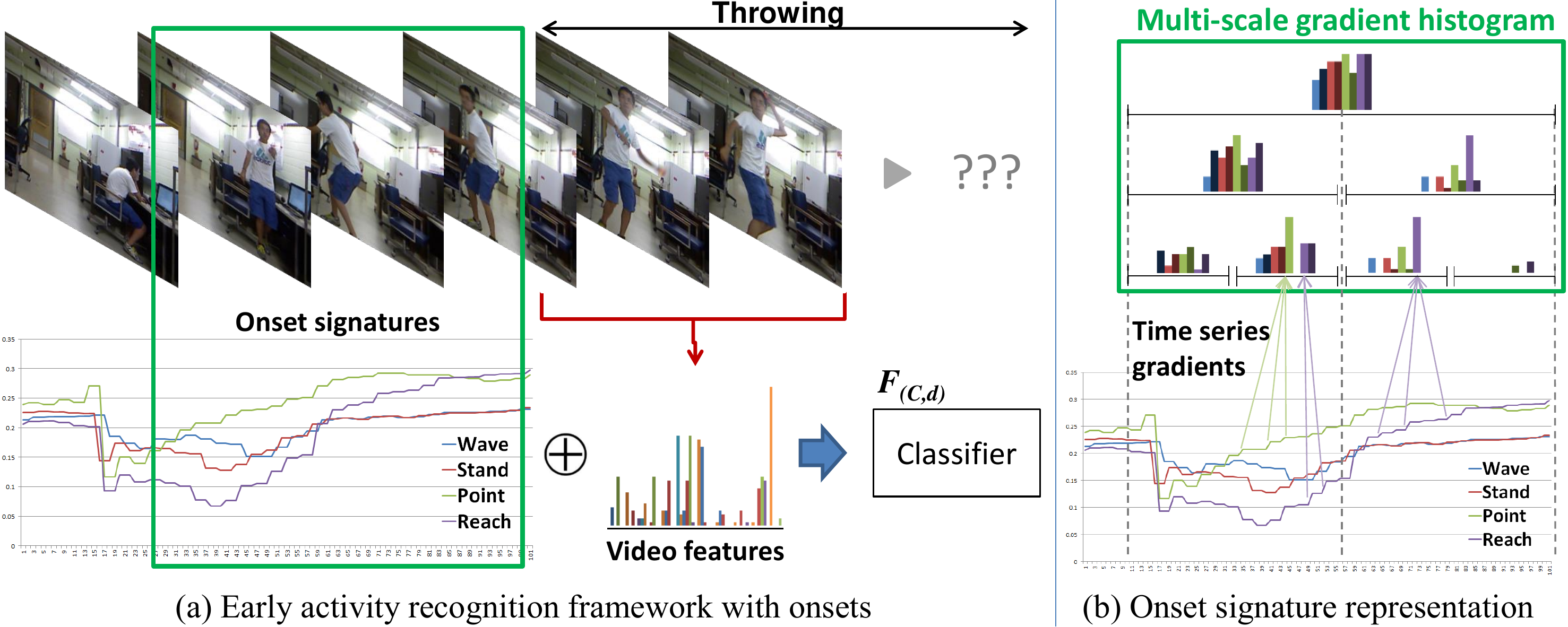}
	}
	\caption{(a) Illustration of the overall concept of our early activity recognition pipeline using onsets. The recognition approach is designed so that it considers both the pre-activity observations (onset signatures) and video features from ongoing videos. (b) Illustration of our onset signature representation, based on multi-scale cascade histogram of time series gradients.}
	\label{fig:recognition}
\end{figure}

In order to enable early detection while addressing the above-mentioned challenges, we introduce the new concept of `onset activities' and `onset signatures' together with our recognition approach to take advantage of them. The idea is to learn weak detectors for subtle short-term activities (i.e., onset activities) which are closely or loosely related to the occurrence of important activities, and make the recognition system to capture activity-activity relations (i.e., $\textbf{A}$) using these onsets. Our approach learns onset patterns leading to the occurrence of a target activity while explicitly considering stochastic nature of onsets, and performs its early recognition by analyzing onset distributions observed before the activity. Figure \ref{fig:recognition} (a) illustrates its overall concept.

\subsection{Onset activities}

We define \emph{onset activities} as subtle activities which (1) occur within a short time duration and (2) do not physically influence the observer (i.e., a wearable camera or a robot), but (3) serve as a direct/indirect cue to infer their following activities. `Standing', `pointing', and `picking up an object' are typical examples of onset activities. These activities themselves do not have strong meaning and they do not influence the observer directly, but they can serve as indicators describing `what activity is likely to follow next'. An example will be the activity of `picking up an object' serving as an onset for `throwing an object'. Another example will be `waving' before `hand shaking'.

Typically, because of the subtle nature of onset activities, their recognition becomes difficult and unreliable. The activities usually contain a small amount of human motion (e.g., only a subtle arm/finger gesture is visible when `pointing'). This makes the detectors for onset activities to become weak classifiers, and prevents the system from directly using the onset recognition results. For instance, average precisions (AP) of our onset activity detection were 0.1 to 0.2 in our dataset. Furthermore, onset activities often have very stochastic nature, implying that onsets are not always observable before activities. Thus, an approach to best utilize these weak detectors is required.


\subsection{Onset signatures}
\label{subsec:signature}

We define \emph{onset signatures} as a set of time series describing onset activity matching results. That is, given a continuous (streaming) video, we measure the similarity between each possible video segment and the onset activity, and record how the similarity is changing over time. The objective is to use these time series as features suggesting future activities. Each onset signature $G^k(t)$ of $k$th onset activity is more specifically computed as:
\begin{equation}
	\begin{aligned}
		G^k(t) = \max_r (1 - D^k([t-r, t]))
	\end{aligned}
\end{equation}
where $r$ is the model duration of the activity, and $D^k([t-r, t])$ is the distance between the model of the $k$th onset activity and the video observation segment $V[t-r, t]$. We use a basic template matching of bag-of-words representations (obtained from a set of training videos $S^k$) as our $D^k$:
\begin{equation}
	\begin{aligned}
		D^k([t_{1}, t_{2}]) = \sum_{i} (m_i^k - v_i[t_{1}, t_{2}]))^2
	\end{aligned}
\end{equation}
where $v_i$ is the $i$th feature value and $m_i^k$ is its mean model value: $m_i^k = \sum_{V^j \in S^k} v_i^j / |S^k|$.


The matching is performed for all $t$ and possible $r$ values, providing us the final $G^k(t)$. The resulting $G^k(t)$ forms a time series, describing how our onset activity detector is responding to the ongoing video observation. We collect $G^k(t)$ from all onset activities and use them as our \emph{onset signature}.


{\flushleft\textbf{Histogram representation of onset signatures:} We design a histogram representation of onset signatures (Figure \ref{fig:recognition} (b)). The idea is to make the system efficiently summarize the previous onset occurrence information from its time series, so that it can use it to infer ongoing/future activities.}

Typical representations of onset signatures are mean and maximum values of a fixed time window (e.g., frames between the current frame and 50 frames before that). However, this is often insufficient due to noisy and weak nature of onset matching (notice that onset recognition relying on peak matching values give us $\sim$0.1 AP), and deeper analysis of time series is necessary. Thus, we construct cascade histograms of time series gradients to represent onset signatures.

Let $||$ denote the concatenation operation of two vectors, $[a_1, \cdots, a_n] ~||~ [b_1, \cdots,$ $b_n] = [a_1, \cdots, a_n, b_1, \cdots, b_n]$. Then, the histogram representation of onset signature $H$ at time $t$ is defined as: $H(t) = H_1(t-b, t) ~||~ H_2(t-b, t) ~||~ \cdots ~||~ H_{|\textbf{A}|}(t-b, t)$, where $H_k(t-b, t)$ is the histogram for the $k$th onset activity computed based on the time interval $[t-b, t]$ with duration $b$ (e.g., 100). $H_k$ is defined more specifically as:
\begin{equation}
	\begin{aligned}
		H_k(t_1, t_2) = & H_k\left(t_1, \frac{t_1+t_2}{2}\right) ~||~ H_k\left(\frac{t_1+t_2}{2}, t_2\right) ~||~ [h_k^+(t_1, t_2), h_k^-(t_1, t_2)]
	\end{aligned}
\end{equation}
where
\begin{equation}
	\small
	\begin{aligned}
		h_k^+(t_1, t_2) &= \left|\{t_1 \leq t \leq t_2 ~|~ G^k(t) - G^k(t-s) > 0 \}\right|,\\
		h_k^-(t_1, t_2) &= \left|\{t_1 \leq t \leq t_2 ~|~ G^k(t) - G^k(t-s) \leq 0 \}\right|.
	\end{aligned}
\end{equation}
Here, $s$ is the step size of gradient computation, and we perform this histogram construction for multiple $s$ scales and concatenate the results.

The above recursive equation hierarchically performs temporal segmentation of the time series (i.e., our onset signatures) into multiple parts, and obtains a histogram of time series gradients corresponding to each of them. That is, our hierarchical histogram is constructed by applying our recursive function until it reaches the level $l$. In our experiments, $l=3$ gave us good results.

The final feature vector representation of the onset signature is constructed as follows, by attaching mean and max values to the histogram:
\begin{equation}
	\begin{aligned}
		x(t) = H(t) ~||~  \left[ \sum_{t'=t-d}^t \frac{G^1(t')}{b}, \cdots,  \sum_{t'=t-d}^t \frac{G^n(t')}{b}\right]~||~ \left[\max(G^1(t')), \cdots,  \max(G^n(t'))\right]
	\end{aligned}
\end{equation}

\subsection{Early detection using onset signature}
\label{subsec:detection}

Based on its video observation $V$ and computed onset signatures $x$, our approach performs early detection of an activity by using a set of binary classifiers. More specifically, we formulate the detector at time $t$ as:
\begin{equation}
	\begin{aligned}
		P(C^t ~|~ V) =& \sum_d \sum_{[t_{1}, t_{2}]} P(C^{[t_{1}, t_{2}]}, d ~|~ V) \propto \sum_d \sum_{[t_{1}, t_{2}]} \sum_{(\textbf{A}, I)} F(C^{[t_1, t_2]}, d, \textbf{A}, I, V) \\
							\approx& \max_{d} \max_{[t_{1}, t_{2}]} \sum_I ~F_{(C, d)} (x(t), V[t_{1}, t]) \cdot e^{w \cdot L_C([t_{1}, t_{2}], I)} \\
	\end{aligned}
\end{equation}
where $t = t_{1} + d \cdot (t_{2} - t_{1})$ and $w$ is the weight. $L_C([t_{1}, t_{2}], I)$ is a typical likelihood of the current activity $C$ occurring at the time interval $[t_{1}, t_{2}]$ under the intention $I$, and we assume a Gaussian distribution of it.

That is, we train one binary classifier for each $F_{(C, d)}$, and perform the recognition of $C$ by considering all possible progress level $d$. A concatenation of the vector describing video features inside the interval (i.e., $V[t_{1}, t]$) and the vector representation of our onset signature (i.e., $x(t)$) serves as an input to these classifiers, and each of the learned classifier $F_{(C, d)}$ measures whether the activity is ongoing at the time $t$ or not. The training of the classifier is performed by providing positive and negative samples of $V$ and $x$ together with their ground truth labels $y$.

The idea is to abstract all previous activity occurrence combination $\textbf{A}$ by focusing on the behaviors of the onset activities that are actually relevant to the current activity $C$. We directly use our onset signatures as features instead of evaluating all possible activity configurations, thereby enabling efficient computations. The classifier will learn to focus on particular onset signature while ignoring irrelevant onset activities. Our approach is able to cope with any types of binary classifiers in principle, and it is able to do it more reliably with classifiers generating scores (e.g., probability estimates). We used support vector machine (SVM)-based probability estimation in our implementation.


Overall computations to recognize a target activity can be performed with $O(n \cdot |d| \cdot R$) at each time step if a binary classifier with a linear amount of computations is used (e.g., SVM), where $n$ is the feature dimension, $|d|$ is the number of possible activity progress levels we consider (we used 10 levels in our experiments), and $R$ is the number of activity durations we consider (this influences possible $t_{1}$).

\section{Experiments}
\label{sec:experiments}

\begin{figure}
	\centering
	\resizebox{1.0\linewidth}{!}{%
	  \includegraphics{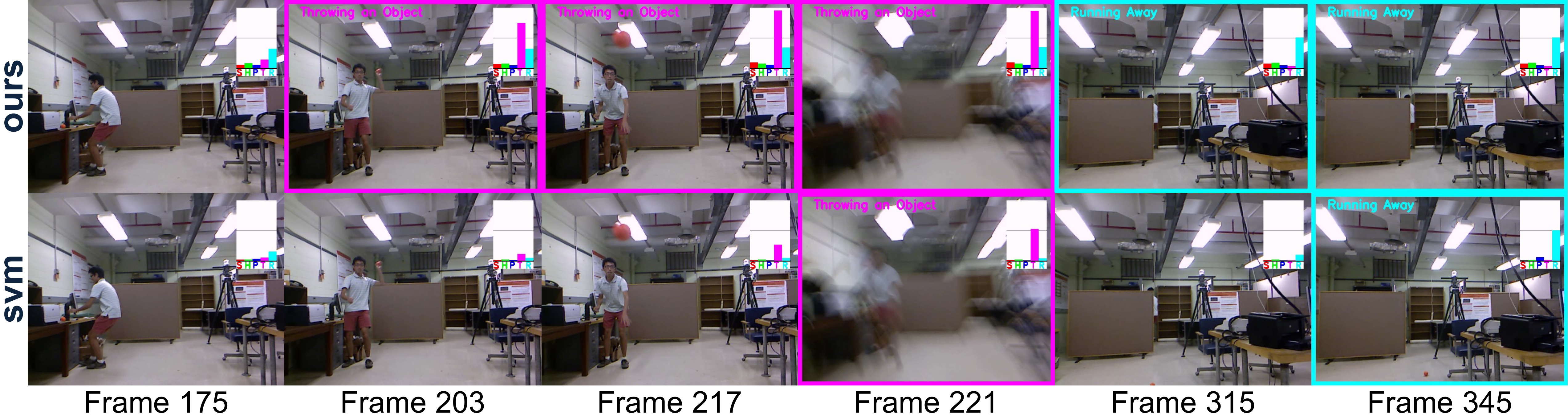}
	}
	\caption{Example result image sequences comparing our early detector (top) with the previous SVM detector using state-of-the-art features (bottom). `Throwing' activity (magenta boxes) and `running away' activity (cyan boxes) are detected.}
	\label{fig:sample-results}
\end{figure}

\subsection{Dataset}

We constructed a new public dataset composed of continuous human activity videos taken with a first-person viewpoint. It is an extension of the previous humanoid-based first-person video dataset \cite{ryoo13} whose videos mostly contain a single activity; our new videos contain a sequence of 2$\sim$6 activities (onsets and interactions). The motivation is that the community has been lacking a public dataset for `early recognition' (i.e., activity prediction): To our knowledge, the only public dataset used for activity prediction (by at least more than 5 previous works) is UT-Interaction \cite{ryoo09iccv}. However, even if we set aside that UT-Interaction is not a first-person video dataset, it has a major limitation: its videos contain activities executed in an random order without any context (e.g., punching and then shaking hands). This is very unnatural, since the actors are following a fixed script without any intention on their own (unlike our new dataset).

Our camera was mounted on top of a humanoid similar to \cite{ryoo13}, and we asked human subjects to perform a series of activities with three different types of intentions: friendly, hostile (i.e., harming), and avoiding. We labeled 9 types of human-observer interactions performed by the actors: 4 types of onset activities and 5 types of activities-of-interest. `Pointing the observer (i.e., the camera)', `reaching an object', `standing up', and `waving to the observer' are the 4 onset activities. `Handshaking with the observer', `hugging the observer', `punching the observer', `throwing an object at the observer', and `running away from the observer' are the 5 interaction-level activities in our dataset. The humanoid moves as it is involved in the interaction (e.g., the camera collapsing as a result of punching). Furthermore, since the operator emulated the humanoid's mobility (rotation and translation) from its behind, camera ego-motion is present in the videos. We also emphasize that the videos were collected at an institution different from the authors' and that the authors had no prior knowledge on these video sequences.


Videos with the resolution of 640*480 with 30fps were used. Also, 320*240 depth videos were collected in addition to the conventional RGB videos, so that the system obtains Kinect-based posture estimation results every frame. The dataset consists of 8 sets, where each set contains continuous videos of human activities being performed by the same subject. The dataset contains a total of 61 continuous videos with more than 180 executions of human activities.

\subsection{Implementation}
\label{subsec:implementation}


We extracted multiple types of state-of-the-arts visual features from first-person videos, including global motion descriptors \cite{ryoo13}, local motion descriptors \cite{laptev05,dollar05}, and human posture descriptors \cite{xia12}. Once features are extracted from raw videos, we clustered these features to obtain standard bag-of-visual-words feature representation while using integral histograms for more efficient computations. Our approach and multiple baseline approaches including the state-of-the-art early recognition approach \cite{ryoo11} were trained/tested/compared using our dataset.

We implemented (1) our approach utilizing onset signatures as well as (2) its simplified version designed to only utilize peak onset activity responses (instead of full onset signatures). In addition, we implemented (3) an extended version of previous state-of-the-art early recognition approach \cite{ryoo11} designed for the 3rd-person videos, and (4) made it to also take advantage of our onset signatures. Furthermore, we implemented several baseline activity detection approaches including (5) a sliding window detector with Bayesian classifiers assuming a Gaussian distribution (i.e., a after-the-fact detection approach), and (6) a SVM-based activity detector using a non-linear kernel (i.e., RBF). All of the above approaches took advantage of the same features vector (i.e., the concatenation of four feature types \cite{laptev05,dollar05,xia12,ryoo13}), which outperformed those using single feature type. We also tested (7) the approach detecting activities solely based on onset signatures (i.e., context-only). 



All these approaches run faster than real-time on a standard desktop PC with our unoptimized C++ implementation (0.0036 sec per frame), except for the adopted feature extraction part.

\begin{figure}
	\centering
	\resizebox{0.75\linewidth}{!}{%
	  \includegraphics{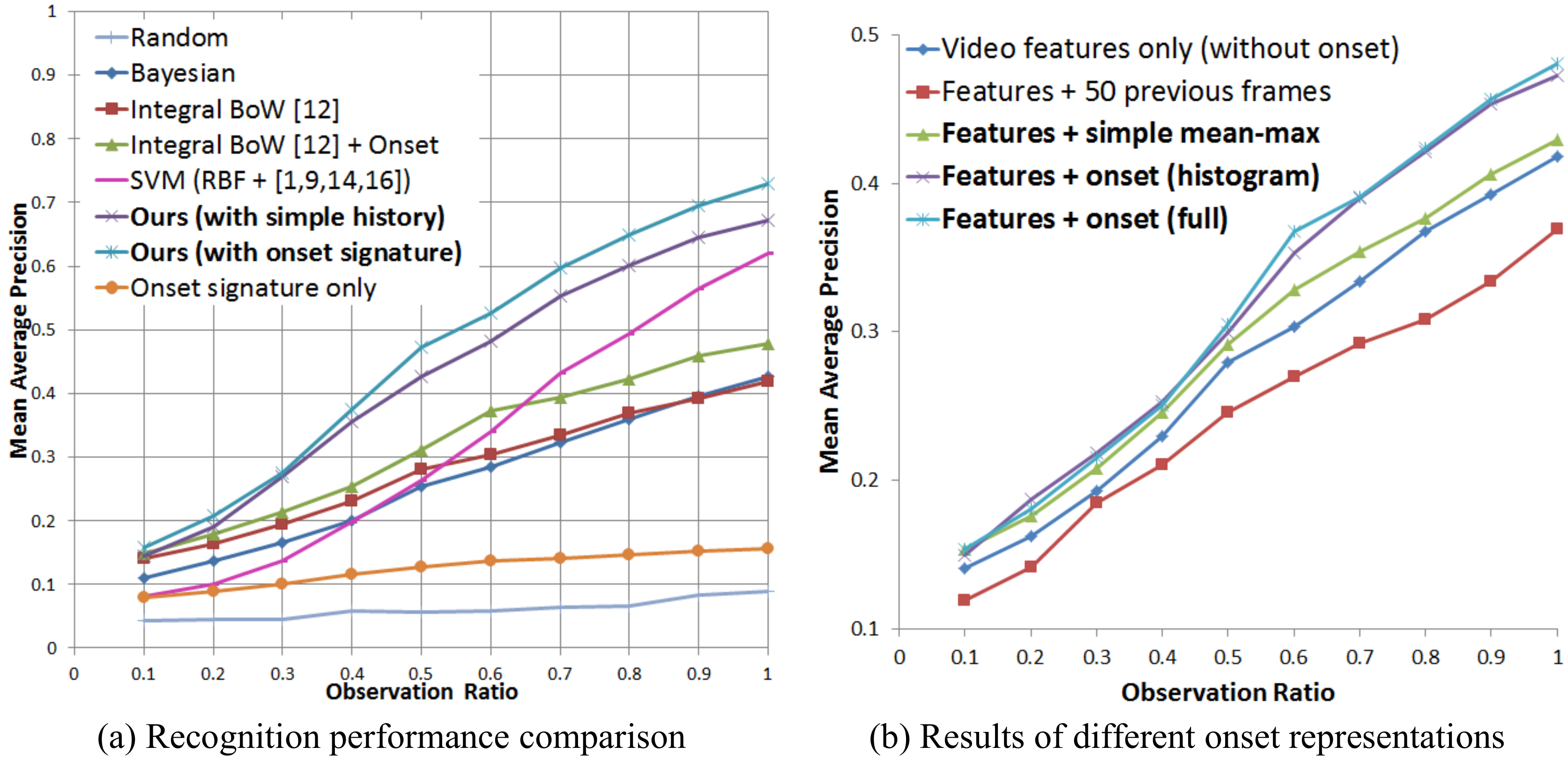}
	}
	\caption{(a) A figure comparing performances of our early detection approaches with previous works and baselines. Mean AP, which is an area under a precision-recall curve, is measured per observation ratio. A higher graph suggests better performance; a higher graph indicates that it `recognizes activities more accurately at the same observation ratio compared to the others' and that `it is able to recognize activities earlier than the others if the same accuracy is assumed'. (b) We compared approaches using different onset representations to illustrate the advantages of our proposed cascade histogram-based onset signature representation, while using the deterministic \cite{ryoo11} as baseline. Both (a) and (b) show benefits of our onsets.}
	\label{fig:results}
\end{figure}

\subsection{Evaluation}

We use leave-one-set-out cross validation (i.e., 8-fold cross validation) for continuous `detection' tasks. Ground truth labels of activity occurrences in videos for both the onset activities and the interaction activities were provided, so that we can take advantage of them for the training (i.e., a supervised learning setting) and testing. At each round, for the testing, our approach computes the probability $P(C^t ~|~ V)$ (which also can be viewed as a confidence score) at every frame $t$. Treating its peak values as detections, we computed precision-recall curves by changing the detection threshold. Average precision (AP) is also obtained from the curve by measuring the area under the PR-curve, and mean AP is computed by averaging APs of all activity classes.

In addition, in order to measure the early detection ability of our approach, we tested our approaches and baselines with multiple different observation ratio settings similar to \cite{ryoo11}. More specifically, activity observation ratio was set from 0.1 to 1.0, and mean AP was measured per observation ratio. An observation ratio specifies the progress level of the activity execution. For example, observation ratio of 0.2 implies that the system was asked to make the detection decision after observing the initial 20$\%$ of the activity (i.e., very early detection), and observation ratio of 1.0 implies that it is a standard after-the-fact detection. Only the detection found before the observation ratio were consider as true positives.




Figure \ref{fig:results} (a) shows the mean AP values of activity detectors measured with various observation ratio settings. We are able to observe that our concept of utilizing onset activities and their signatures is benefiting the system greatly. Mean APs of our approach using onset were constantly higher by 0.1$\sim$0.2 compared to the SVM classifier using state-of-the-art features, achieving the same AP much earlier. For instance, in order to obtain the mean AP of 0.5, our early detector with onsets requires 55$\%$ observation while the SVM requires more than 80$\%$. This gap can also be observed for integral bag-of-words with and without onsets. Figure \ref{fig:sample-results} shows example images of these detection results.



\begin{figure}
	\centering
	\begin{minipage}[t]{0.57\linewidth}
		\resizebox{1.0\linewidth}{!}{
			\includegraphics{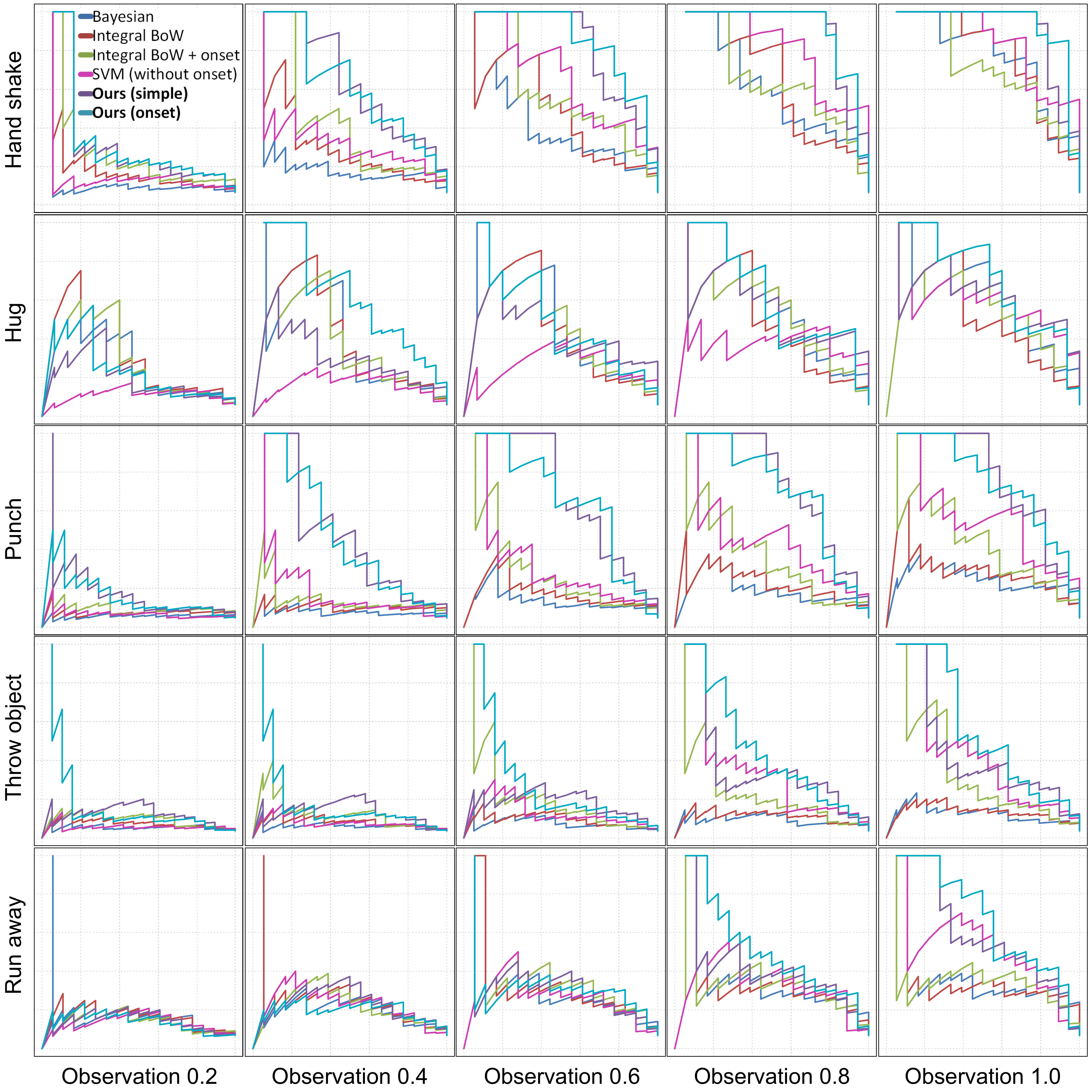}
		}
	\end{minipage}\hfill
	\begin{minipage}[b]{0.43\linewidth}
		\resizebox{1.0\linewidth}{!}{
			\includegraphics{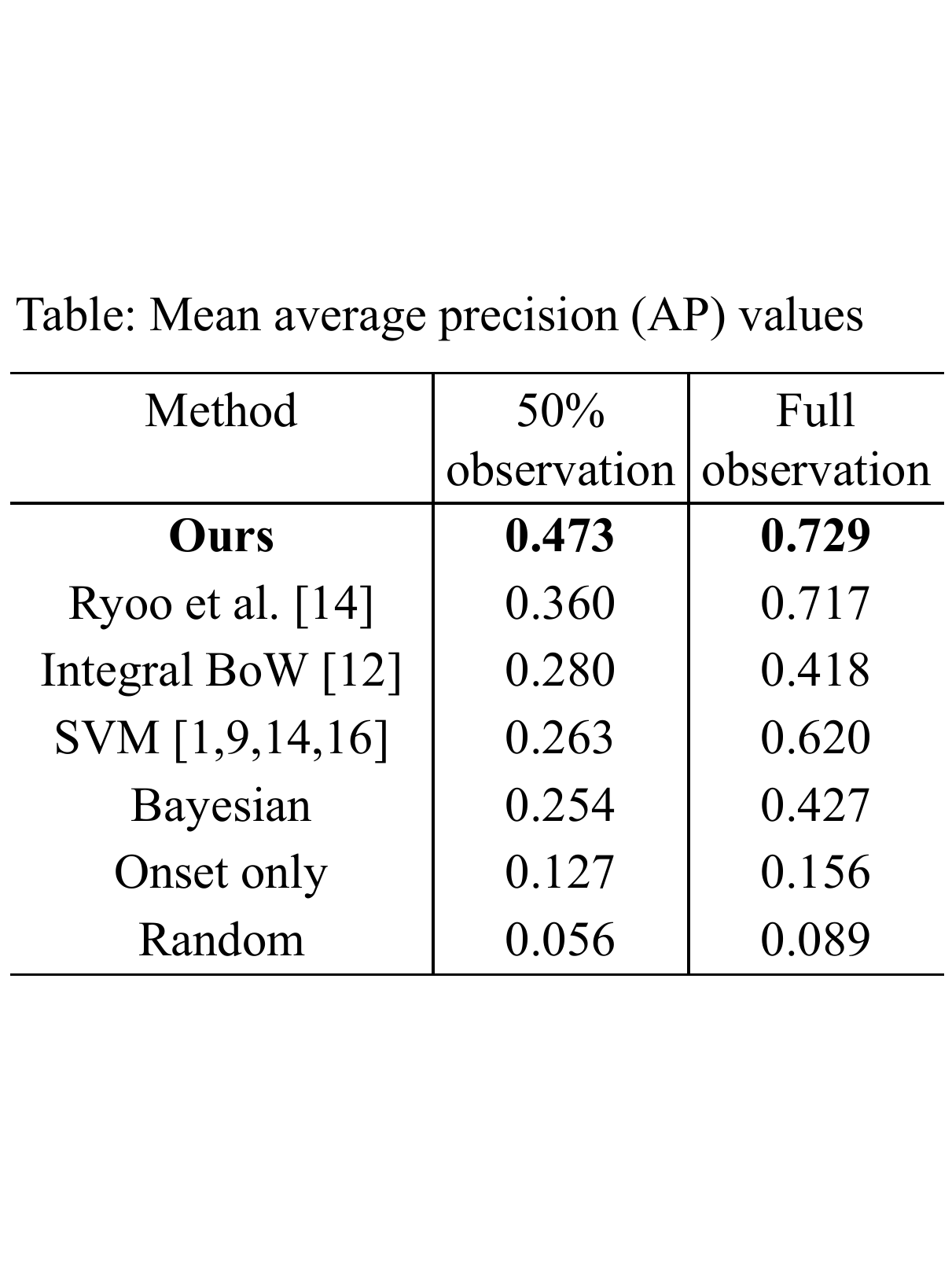}	
		}
	\end{minipage}
	\caption{Precision-recall curves of each activity per observation ratio setting (left) and a table comparing accuracies of state-of-the-arts (right). X axis [0$\sim$1.0] of all graphs is `recall' and Y axis [0$\sim$1.0] is `precision'. Our approach with onset showed the best performance in all cases. Using our onset signature (light blue) particularly showed a huge performance increase over SVM (pink).}
	\label{fig:pr}
\end{figure}

Figure \ref{fig:pr} illustrates actual PR curves of the approaches. Early recognition approaches with our onset signatures particularly performed well on the activity of `throwing an object', since it very often had a clear onset activity: `reaching the object'. Our approach also performed well for `hugging' and `shaking' (relying on the existence of the onset activity `waving' and the absence of `reaching' or `pointing'), and detected `punching' earlier than those not using onsets.

We also conducted an additional experiment to investigate advantages of our cascade histogram-based onset signature representation. We compared the performance of our onset representation with various other onset representations, including (i) the approach adding video features obtained 1$\sim$50 frames prior to the activity in addition to those from the activity's actual video segment, (ii) the approach using a simple onset representation of mean and max values, (iii) the approach only using our histogram-based onset representation, and (iv) our final onset representation composed of histogram + mean and max. Figure \ref{fig:results} (b) illustrates the result. It clearly shows that our onset signature representation effectively captures previous video information. Particularly, we are able to observe that simply adding 50 frames prior to the time interval is only confusing the system. Integral BoW was used as the base classifier in this experiment. 


\textbf{Conclusion:} This paper presented a methodology for early recognition of human activities. Early recognition ability is very essential for first-person vision systems which are required to function in real-world environments in real-time, and this paper investigates such scenarios for the first time. We formulated the early recognition problem to consider pre-activity observations, and presented an efficient new approach that utilizes the concept of onsets. Experimental results confirmed that our formulation enables superior early recognition performance to previous conventional approaches.

{\flushleft\textbf{Acknowledgment:} The research described in this paper was carried out at the Jet Propulsion Laboratory, California Institute of Technology, under a contract with the National Aeronautics and Space Administration. This research was sponsored by the Army Research Laboratory and was accomplished under Cooperative Agreement Number W911NF-10-2-0016.}

\bibliographystyle{ieee}
\bibliography{hri}

\end{document}